\documentclass{article}

\usepackage[english]{babel}

\usepackage[letterpaper,top=2cm,bottom=2cm,left=3cm,right=3cm,marginparwidth=1.75cm]{geometry}

\usepackage{amsmath}
\usepackage{graphicx}
\usepackage[colorlinks=true, allcolors=blue, breaklinks=true]{hyperref}
\usepackage{breakcites}
\twocolumn 

\title{Milimili. Collecting Parallel Data via Crowdsourcing}
\author{Alexander Antonov}

\begin{document}
\maketitle

\begin{abstract}
We present a methodology for gathering a parallel corpus through crowdsourcing, which is more cost-effective than hiring professional translators, albeit at the expense of quality. Additionally, we have made available experimental parallel data collected for Chechen-Russian and Fula-English language pairs.
\end{abstract}

\section{Introduction}

NMT \cite{vaswani2017attention} has made significant advancements in recent years, particularly in enhancing the quality of translations for low-resource languages. However, the relative scarcity of data on the internet makes creating parallel data from scratch quite challenging.

Large multilingual datasets like NLLB \cite{nllbteam2022language} and NTREX \cite{federmann-etal-2022-ntrex}, which have been collected with the help of professional translators, can be quite expensive. In response to this, we conducted an experiment to ascertain if it's feasible to create affordable parallel data with crowd assistance. We included Chechen-Russian and Fula-English language pairs in our research—languages diverse enough to allow evaluating different scenarios.

All the data is available on GitHub\footnote{\href{https://github.com/AlAntonov/milimili}{https://github.com/AlAntonov/milimili}}.

\section{Crowd Settings}

All the experiments were conducted on the Toloka crowd platform\footnote{\href{https://toloka.ai}{https://toloka.ai}} \cite{pavlichenko2021crowdspeech}, where anyone can register and carry out various data labeling tasks.

In our case, an additional prerequisite for task participation was the knowledge of two languages. We did not create specific language tests. It was sufficient for the user to indicate their language proficiency in the settings.

The complete pipeline consisted of two tasks: translation and quality control. Additionally, automatic control was employed.

\subsection{Translation Task}

The translation task was straightforward where we displayed a source sentence and requested a translation into the target language. The instruction was as simple as "Translate the sentence from SRCLANG to TGTLANG".

Only workers who had indicated in their settings that they were proficient in both SRCLANG and TGTLANG were allowed to undertake this task.

In total, we had four translation tasks: from Chechen to Russian, from Russian to Chechen, from Fula to English, and from English to Fula.

\subsection{Quality Control Task}

Translated sentences were sent to the quality\\control task where we asked workers to evaluate whether the translation of the source sentence was good or bad. To ensure the accuracy of the evaluation, we gave the same translation to three workers. The majority vote was selected.

For the quality control task, we had created an additional exam, so performers had to prove their language proficiency. However, the exam was fairly easy. We had ten sentences. Five sentence pairs were correct translations, mined from various parallel resources. The other five sentence pairs were incorrect translations. Two were translations of different sentences, one was a translation into a different language, and two were word-for-word translations collected from an online dictionary, specifically Glosbe\footnote{\href{https://glosbe.com}{https://glosbe.com}}.

In total, we had the same four tasks for different language directions.

\subsection{Automatic Control}

Certain bad translations could be detected automatically, so we didn't need to send all translated sentences for human verification.

To verify if the translation was in the appropriate language, we used a language detector, specifically Fasttext \cite{joulin2016fasttextzip}.

We also conducted a length relation check between the source and translated sentences. If the difference exceeded three times, we rejected the translation. 

The entire pipeline was prepared so that it could run automatically. However, we carried out manual verification from time to time to check the quality.

\subsection{Crowd Quality}

We utilized basic crowdsourcing tools to detect poor quality, such as excessively fast responses.

Generally, our method is not recommended for language pairs that are already represented in machine translation systems, as people began using machine translation to complete the task.

Furthermore, we want to highlight that the quality of the translated sentences cannot be compared with professional translation. Therefore, we recommend using crowd-translated tasks only in training sets, excluding test sets.

\subsection{Price}

Addressing the problem of low resource languages can entail significant costs. Let's consider a scenario where there are approximately 7,000 languages, each needing a million parallel texts. This totals 7 billion sentences. If we estimate the translation of one sentence at around one dollar, the total expense would rise to 7 billion dollars. Clearly, that is a quite substantial amount.

We paid \$0.02 for each translation and \$0.01 for every set of 10 verifications.

In fact, the total expenditure for all the experiments discussed in this context amounted to approximately \$100.

\section{Data}

As mentioned above, we used source sentences in four languages. We decided to utilize two different language pairs to better understand scalability. One language in each pair was a low-resource language, and the other was a high-resource language. It's worth noting that the low-resource language could potentially link to other languages throw high-resource language.

\subsection{Chechen}

We had taken all the source sentences in Chechen from Wikipedia\footnote{\href{https://ce.wikipedia.org}{https://ce.wikipedia.org}}. From the raw data, we had discarded the template sentences for frequently repeated words. Additionally, we had filtered sentences in other languages using Fasttext \cite{joulin2016fasttextzip}.

\subsection{Russian}

For the Russian language, we used the data from the WMT21 test set \cite{akhbardeh-etal-2021-findings}, which was presented at the News Task.

\subsection{Fula}

All the source sentences in Fula were taken from Wikipedia\footnote{\href{https://ff.wikipedia.org}{https://ff.wikipedia.org}}, in the same way as the Chechen language. It's worth noting that even monolingual data for Fula was quite difficult to find on the Internet. One more important point about Fula is that it has several dialects which are significantly different from each other. We attempted to work with Nigerian Fulfulde but didn't impose strict restrictions.

\subsection{English}

For English, we also used the WMT21 News Task test set \cite{akhbardeh-etal-2021-findings}.

\section{Results and Discussion}

\subsection{Results}
Table~\ref{tab:widgets} shows the main results of the experiment. The first row is about how many sentences were initially translated. The second row shows how many of them were verified. In the case of Fula, the quantity is less than the number of translated sentences since we don't have enough users who passed our exam. The final row tells us how many sentences were marked as good and hence were included in the corpus.

All the data is available on GitHub.

\begin{table*}[t]
\centering
\begin{tabular}{|l|r|r|r|r|r|}
\hline
& Total & fuv-eng & eng-fuv & che-rus & rus-che \\
\hline
Translated & 1627 & 220 & 311 & 491 & 605 \\
Verified & 1470 & 88 & 286 & 491 & 605 \\
In corpus & 1078 & 53 & 176 & 380 & 469 \\
\hline
\end{tabular}
\caption{\label{tab:widgets}The number of sentences collected at each step of the process, both in total and for each language direction.}
\end{table*}

\subsection{Scaling to Other Languages}

Much to our chagrin, we must acknowledge that scaling to other languages is exceedingly difficult due to the scarcity of users who are bilingual and willing to contribute to a crowdsourcing platform. Although we can confidently declare that the proposed method is able to gather sufficient data for training in the Chechen and Russian languages, the overall experiment cannot be deemed a success. This is because we encountered a shortage of contributors for the Fula and English pair. Other language pairs also faced similar challenges.

We believe that the future expansion of crowdsourcing platforms could potentially help ameliorate this issue.

\subsection{Corpus quality}

At the moment, we do not have any proof that the quality of the corpus is good enough to be used in machine translation model training. However, we believe that it is at least as good as automatically mined parallel corpora.

\bibliographystyle{apalike}
\bibliography{sample}

\begin{thebibliography}{}

\bibitem[Akhbardeh et~al., 2021]{akhbardeh-etal-2021-findings}
Akhbardeh, F., Arkhangorodsky, A., Biesialska, M., Bojar, O., Chatterjee, R.,
  Chaudhary, V., Costa-jussa, M.~R., Espa{\~n}a-Bonet, C., Fan, A., Federmann,
  C., Freitag, M., Graham, Y., Grundkiewicz, R., Haddow, B., Harter, L.,
  Heafield, K., Homan, C., Huck, M., Amponsah-Kaakyire, K., Kasai, J.,
  Khashabi, D., Knight, K., Kocmi, T., Koehn, P., Lourie, N., Monz, C.,
  Morishita, M., Nagata, M., Nagesh, A., Nakazawa, T., Negri, M., Pal, S.,
  Tapo, A.~A., Turchi, M., Vydrin, V., and Zampieri, M. (2021).
\newblock Findings of the 2021 conference on machine translation ({WMT}21).
\newblock In {\em Proceedings of the Sixth Conference on Machine Translation},
  pages 1--88, Online. Association for Computational Linguistics.

\bibitem[Federmann et~al., 2022]{federmann-etal-2022-ntrex}
Federmann, C., Kocmi, T., and Xin, Y. (2022).
\newblock {NTREX}-128 {--} news test references for {MT} evaluation of 128
  languages.
\newblock In {\em Proceedings of the First Workshop on Scaling Up Multilingual
  Evaluation}, pages 21--24, Online. Association for Computational Linguistics.

\bibitem[Joulin et~al., 2016]{joulin2016fasttextzip}
Joulin, A., Grave, E., Bojanowski, P., Douze, M., Jégou, H., and Mikolov, T.
  (2016).
\newblock Fasttext.zip: Compressing text classification models.

\bibitem[Pavlichenko et~al., 2021]{pavlichenko2021crowdspeech}
Pavlichenko, N., Stelmakh, I., and Ustalov, D. (2021).
\newblock Crowdspeech and voxdiy: Benchmark datasets for crowdsourced audio
  transcription.

\bibitem[Team et~al., 2022]{nllbteam2022language}
Team, N., Costa-jussà, M.~R., Cross, J., Çelebi, O., Elbayad, M., Heafield,
  K., Heffernan, K., Kalbassi, E., Lam, J., Licht, D., Maillard, J., Sun, A.,
  Wang, S., Wenzek, G., Youngblood, A., Akula, B., Barrault, L., Gonzalez,
  G.~M., Hansanti, P., Hoffman, J., Jarrett, S., Sadagopan, K.~R., Rowe, D.,
  Spruit, S., Tran, C., Andrews, P., Ayan, N.~F., Bhosale, S., Edunov, S., Fan,
  A., Gao, C., Goswami, V., Guzmán, F., Koehn, P., Mourachko, A., Ropers, C.,
  Saleem, S., Schwenk, H., and Wang, J. (2022).
\newblock No language left behind: Scaling human-centered machine translation.

\bibitem[Vaswani et~al., 2017]{vaswani2017attention}
Vaswani, A., Shazeer, N., Parmar, N., Uszkoreit, J., Jones, L., Gomez, A.~N.,
  Kaiser, L., and Polosukhin, I. (2017).
\newblock Attention is all you need.

\end{thebibliography}

\end{document}